\documentclass{article}
\usepackage{spconf,amsmath,graphicx}
\usepackage{subcaption}
\usepackage{hhline}
\usepackage{algorithm}
\usepackage{algorithmic}
\usepackage{amssymb}
\usepackage{amsmath}    
\usepackage{multirow}
\usepackage{color}
\usepackage{booktabs}
\usepackage{balance}
\usepackage[utf8]{inputenc}
\usepackage{tikz}
\usetikzlibrary{positioning, fit, arrows.meta, shapes}

\usepackage{graphicx}
\usepackage{chngpage}
\usepackage{calc}
\usepackage{pifont}
\usepackage{cite}

\newcommand{\cmark}{\ding{51}}%
\newcommand{\xmark}{\ding{55}}%

\copyrightnotice{978-1-6654-7189-3/22/\$31.00~\copyright2023 IEEE}

\title{Accelerator-Aware Training for Transducer-based Speech Recognition}
\name{\parbox{\linewidth}{\centering Suhaila M. Shakiah, Rupak Vignesh Swaminathan, Hieu Duy Nguyen, Raviteja Chinta$^\dagger$, Tariq Afzal$^\dagger$, Nathan Susanj, Athanasios Mouchtaris, Grant P. Strimel, Ariya Rastrow}}
\address{Alexa Machine Learning, Amazon, USA; $^\dagger$Hardware Compute Group, Amazon, USA}
\begin{document}
\maketitle
\begin{abstract}
Machine learning model weights and activations are represented in full-precision during training. This leads to performance degradation in runtime when deployed on neural network accelerator (NNA) chips, which leverage highly parallelized fixed-point arithmetic to improve runtime memory and latency. In this work, we replicate the NNA operators during the training phase, accounting for the degradation due to low-precision inference on the NNA in back-propagation. Our proposed method efficiently emulates NNA operations, thus foregoing the need to transfer quantization error-prone data to the Central Processing Unit (CPU), ultimately reducing the user perceived latency (UPL). We apply our approach to Recurrent Neural Network-Transducer (RNN-T), an attractive architecture for on-device streaming speech recognition tasks. We train and evaluate models on 270K hours of English data and show a 5-7\% improvement in engine latency while saving up to 10\% relative degradation in WER.
\end{abstract}
\begin{keywords}
Accelerator-aware training, model compression, automatic speech recognition (ASR), recurrent neural network transducer (RNN-T).
\end{keywords}
\section{Introduction}
\label{sec:intro}
\begin{figure}[!t]
	\centering
	\includegraphics[width=0.8\linewidth]{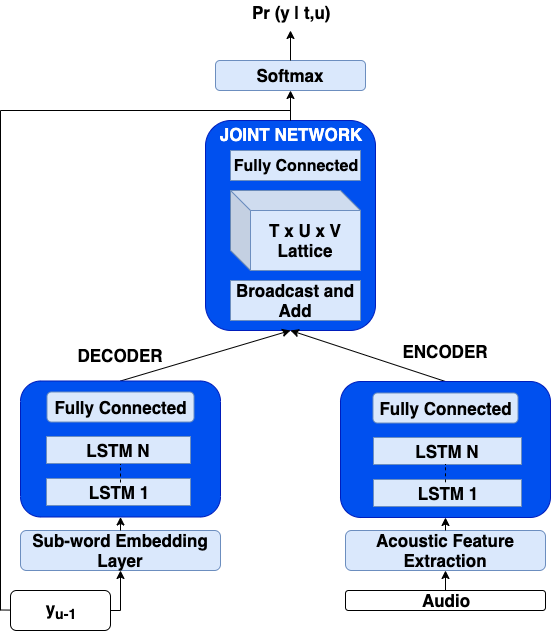}
	\caption{Model diagram of Recurrent Neural Network Transducer (RNN-T).}
	\label{fig:rnnt}
	\vspace*{-1mm}
\end{figure}

The task of transcribing an audio to the corresponding text transcriptions constitutes the Automatic Speech Recognition (ASR) component of voice assistants such as Alexa, Google Home or Siri. ASR solutions have evolved from traditional hybrid Deep Neural Network (DNN) - Hidden Markov Model (HMM) systems to modern end-to-end neural architectures including various transducer-based systems \cite{lu2020exploring,Wang_2020,8462506,radfar2022convrnnt,shangguan2019optimizing}. While many of these approaches have demonstrated high accuracy, an important differentiation comes from their streaming capability, which reduces the user's perceived latency (UPL). A streaming ASR system is able to start transcribing the audio even before the user has finished speaking an utterance, i.e., the system does not require future context to inform the transcription results. Among streaming architectures, the Recurrent Neural Network-Transducer (RNN-T) model (see Figure \ref{fig:rnnt}) \cite{Graves2012} stands out as an all-neural, end-to-end (E2E) method with both low latency and high accuracy, and it is widely adopted in modern speech recognition systems. 

Apart from the current trend of moving to E2E architectures for machine learning (ML) applications, leading ML solution providers also utilize on-device processing to improve user experience and reduce UPL. As a result, hardware consisting of NNA chips have been gradually deployed to support on-device computer vision, natural language understanding (NLU), and ASR tasks. However, this comes with additional challenges when moving from floating point to fixed point operations supported by NNAs. Since computers are finite state machines, real numbers are represented and manipulated in floating point format in computer memory. Floating point numbers are characterized by a mantissa, an exponent and the sign bit, which enable the representation of a wide range of values with a floating decimal point. For example, a 32-bit floating point (FP-32) number can represent a total of $2^{32}$ unique values with a higher precision than their fixed point counterparts. In the context of performing machine learning on the edge, floating point computations are time consuming and memory expensive, especially for deep learning models used for E2E speech recognition which involve millions (and sometimes billions) of multiply-and-accumulate (MAC) operations for a single inference cycle. To address the time and memory complexity, neural accelerator chips thus employ fixed point operations, in which each value is normally represented by a reduced number of bits.

\begin{figure}[!hptb]
	\centering
	\includegraphics[width=0.8\linewidth]{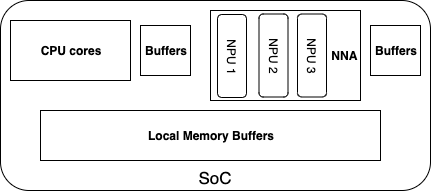}
	\caption{A representative SoC showing CPU and NPU cores with local memory and data buffers}
	\label{fig:soc}
	\vspace*{-1mm}
\end{figure}

It is worth noting that neural ML models are highly sensitive to such reductions in the precision of weights and activations, and this effect is even more pronounced in recurrent architectures since the errors accumulate across multiple time steps. To address these problems, while also being cognizant of the low latency, power and memory requirements for on-device systems, the on-device chips adopt a hybrid architecture which includes general-purpose central processing units (CPUs) as well as specialized neural processing unit (NPU) cores, integrated into a single System-On-Chip (SoC) design (see Figure \ref{fig:soc}). In order to acheive the best performance, certain computations are performed in highly efficient NPUs whereas others requiring a higher precision are computed on CPUs. This trade-off results in additional on-chip computing and data transfer latency between the NPU and CPU, creating bottlenecks during inference. For example, on-device ASR models need to use the CPU to compute tanh and sigmoid activations, which are not only quantization error-prone on NPU, but also computation and memory intensive. In this work, we aim to reduce the latency incurred due to moving data between the NPU and CPU to perform non-linear activations. Using an ASR task and RNN-T architecture as a proof of concept, we show that we are able to improve model inference speed by 20\% on-device with negligible, less than 1\% relative, accuracy degradation by performing accelerator-aware training (AAT), thus making the models more robust to the NNA and the hardware activation functions. 

The rest of the paper is structured as follows. Section \ref{sec:related} describes the related work for latency reduction in RNN-T speech models, followed by an overview of quantization in neural networks and NNAs in Section \ref{sec:overview}, as well as details of accelerator quantization schemes in Section \ref{sec:nna}. Section \ref{sec:AAT} describes our proposed accelerator-aware training technique. We then dive into the experimental details in Section \ref{sec:setup} and the performance and latency results in Section \ref{sec:results}. Section \ref{sec:concs} concludes our paper with some remarks on the advantages of the proposed approach.

\section{Related Work}
\label{sec:related}
There is a large body of work focussing on improving the Word Error Rate (WER) and runtime latency for on-device RNN-T models. Architectural modifications to recurrent neural networks, such as CIFG-LSTM \cite{greff2016lstm} and Simple Recurrent Units \cite{lei2017simple} have been used in RNN-T saving 30-40\% compute while having negligible impact on the WER. Knowledge distillation techniques specific to RNN-T have also been studied \cite{panchapagesan2021efficient,swaminathan2021codert,nagaraja2021collaborative}.  Employing sparsity-based pruning on the weight matrices of LSTMs have been studied for both structured and unstructured sparse matrices \cite{shangguan2019optimizing, zhen2021sparsification}. In \cite{yu2021fastemit}, the authors proposed an additional regularization term in the RNN-T loss to penalize blank token prediction so that the model emits the labels faster, leading to a reduction in latency while maintaining the WER. In \cite{9414652}, the authors introduced a bifocal encoder architecture for RNN-T to improve streaming latency, where the low entropy wakeword segment of an utterance is processed by a small encoder allowing the larger encoder to catch up during decoding of the rest of the utterance as the frames become available. As a more general approach to switching between encoders of different compute capacity, \cite{macoskey2021amortized} proposed an arbitrator network that could dynamically choose the encoder network on a per-frame basis.

In addition to modified architectures and loss functions, there have also been many advances in training methods that emulate inference; quantization-aware training (QAT) and sparse pruning methods have been proposed to improve on-device runtime latency without hurting accuracy \cite{NguyenAle2020,alvarez2016efficient,jacob2017quantization,han2016deep,ZhuGup2018,ZhenNguyen2021}. In \cite{kang2019accelerator}, the authors propose an accelerator-aware neural design where the architecture search space is explored to opimize performance on NNA. More relevant to this work is \cite{li2021quantization}, in which the authors describe a quantization strategy for RNN-T based speech recognition systems using 16-bit activations. This work focuses on accelerators that uses 8-bit activations, demonstrating on-par performance with unquantized baselines even for models with smaller number of parameters, which are more susceptible to quantization.

\section{Overview of Quantization}
\label{sec:overview}

In this section, we provide a brief overview of the quantization schemes, notations and definitions used in this paper. 
Quantization is the process of converting floating point values to a smaller set of discrete fixed point values, effectively reducing the number of bits used to represent the numbers. We will use the $Q$-notation to define the parameters of a signed fixed point number. A fixed point number denoted by  $Q_{m.n}$  has the following properties:
\begin{itemize}
    \item Total bit width is $m + n$, including $m$ integer and sign bits plus $n$ fractional bits.
    \item Consider a signed representation, i.e., the most significant bit (MSB) is 1 for negative values. The minimum possible fractional value that can be represented, $f_{min}$,  is  $-2^{m-1}$, while the maximum representable fractional value, $f_{max}$, is $2^{m-1} - 2^{-n}$.
    \item The resolution is $2^{-n}$, yielding a maximum quantization error of $2^{-(n+1)}$ between a number and its quantized counterpart.
\end{itemize}

For example, $Q3.2$ has 5 bits in total, 3 integer bits with the MSB indicating the sign bit, 2 fractional bits and can represent fractional numbers in the interval [-4, 3.75] with a resolution of $0.25$, thus able to represent a total of $2^{5}$ unique values. 

\subsection{Static Quantization}
In static quantization, the input value $R_i$, is first clipped to be within the quantizable integer range, thus accumulating clipping error. Afterwards, the resulting FP-32 value is scaled and rounded to the nearest integer, which is then scaled back to its floating point equivalent. This ensures that the numbers, during inference on-device, will have the lowest quantization (rounding) errors. For a $Q_{m.n}$ fixed point quantization scheme, the quantized floating point equivalent, $R_q$, of an input FP-32 value $R_i$ is given by,

\begin{equation}
\label{eq:quantdequant}
R_q =  R \left(  C \left(R_i, f_{min}, f_{max} \right) 2^n\right)  2^{-n}
\end{equation}
in which $R$ denotes $Round$ function, which rounds the values to the nearest integer or towards zero depending on the implementation and $C$ denotes $Clip$ function which clips the values at $f_{min}$ and $f_{max}$. In this work, we round the weights to their nearest value and round the inputs and hidden states towards zero as implemented on the accelerator.

\subsection{Dynamic Quantization}
In dynamic quantization, there is an additional degree of freedom, in which the input values can be scaled with a scaling factor, so that the range fits into the quantizable range of integers. This scaling factor needs to be a power of 2 and is calculated dynamically for every sample and time step for the NNA inference. Simply put, the scaling factor, $S$, is the smallest power of 2 that scales the incoming tensor and fits all values within the quantizable range of the given target $Qm.n$ scheme. A clipping error is incurred if a larger scaling factor is required. After the quantization, the outputs are scaled back up. The modified equation is as follows:
\begin{equation}
\label{eq:quantdequant_dyn}
R_q = S \times R \left(  C \left( \frac{R_i} {S}, f_{min}, f_{max} \right) 2^n\right)  2^{-n}
\end{equation}
where $S \in \{1, 2, 4, 8, 16\} $,  such that $ f_{min} \leq \frac{R_i} {S}  \leq f_{max}  $.

\section{Neural Network Accelerator}
\label{sec:nna}

In this section, we briefly discuss how neural network weights and activations are quantized in our NNA experiments.
The number of bits for the quantization of weights and activations, the type of quantizations performed, supported layers, the specific data paths between the various on-chip components, etc., differ from one chip version to another. In our experiments, we use the following schemes: 
\begin{itemize}
\item \textbf{LSTM Cells}: The hidden states are statically quantized to $Q1.7$ in all LSTM cells in the encoder and decoder.
\item The inputs to the first LSTM layer in the encoder and decoder are dynamically quantized to  $Q1.7$ (because it directly follows a CPU data path, hence will not incur additional latencies to compute dynamic scaling factors). The inputs to all other LSTM layers in the encoder and decoder are statically quantized to  $Q1.7$
\item The sigmoid and tanh activations for calculating the gate values $i$,  $f$, $g$ and $o$ in the LSTM cells are non-uniformly quantized 8-bit values as shown in Figs. \ref{fig:sigmoid} and \ref{fig:tanh}. As sigmoid and tanh are non-linear, non-uniform quantizations are more suitable to reduce the means-squared error (MSE) than uniform quantization.
     	
\item \textbf{Dense Layers}: The inputs to all dense layers in the encoder, decoder and joint network are dynamically quantized to $Q1.7$.
\item \textbf{Weights}: All weights in the embedding layer, encoder, decoder and joint network are statically quantized to $Q1.7$
\end{itemize} 

\subsection{Hidden States and Inputs Quantization}
\begin{figure*}[hptb!]
\centering

\begin{subfigure}{0.36\textwidth}
\includegraphics[width=0.99\textwidth]{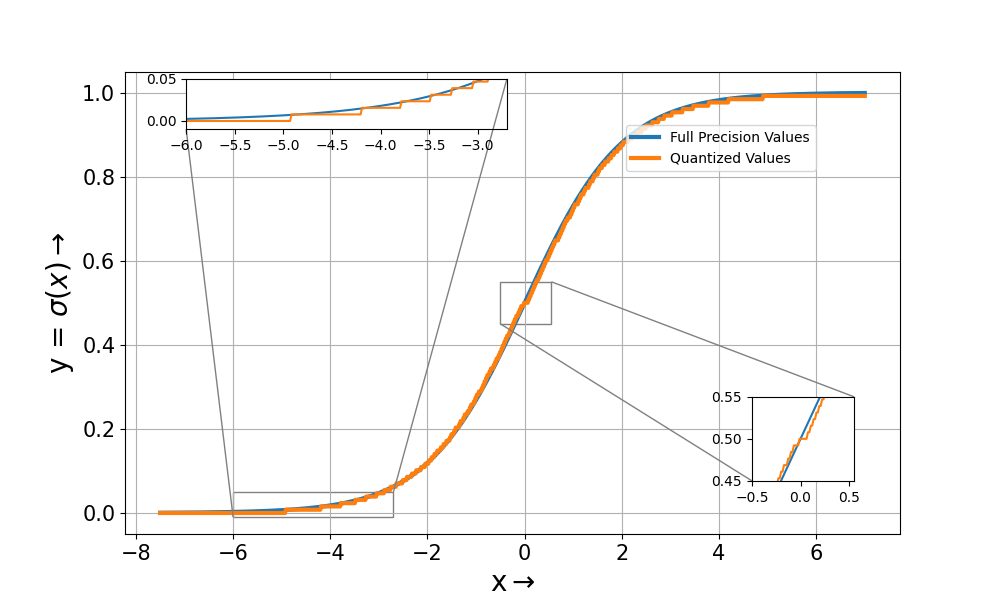}
\caption{Sigmoid Function}
\label{fig:sigmoid}
\end{subfigure}%
\begin{subfigure}{0.36\textwidth}
\includegraphics[width=0.99\textwidth]{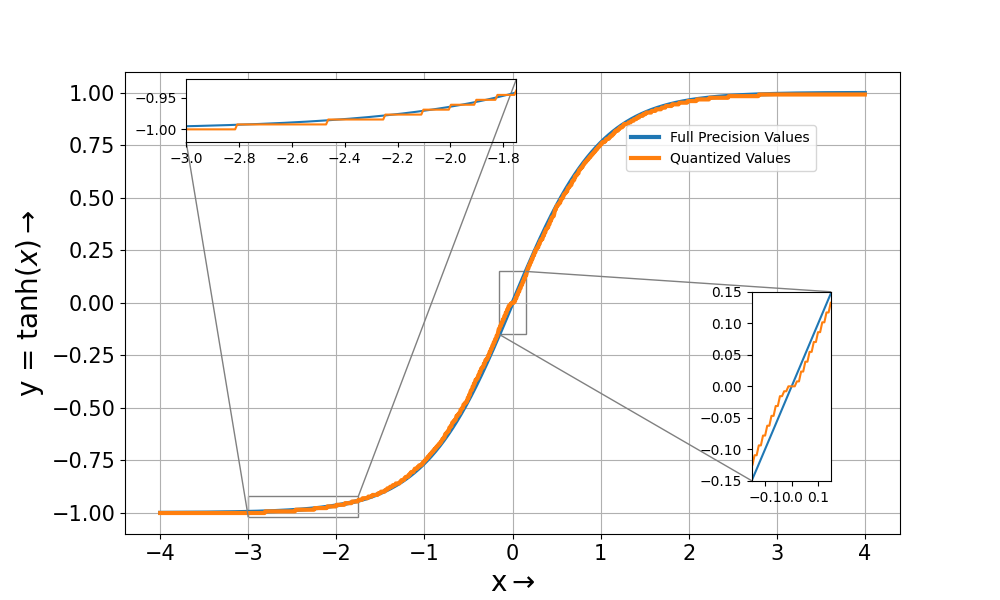}
\caption{Tanh Function}
\label{fig:tanh}
\end{subfigure}%
\begin{subfigure}{0.3\textwidth}
\vspace{0.5cm}
\includegraphics[width=0.9\textwidth]{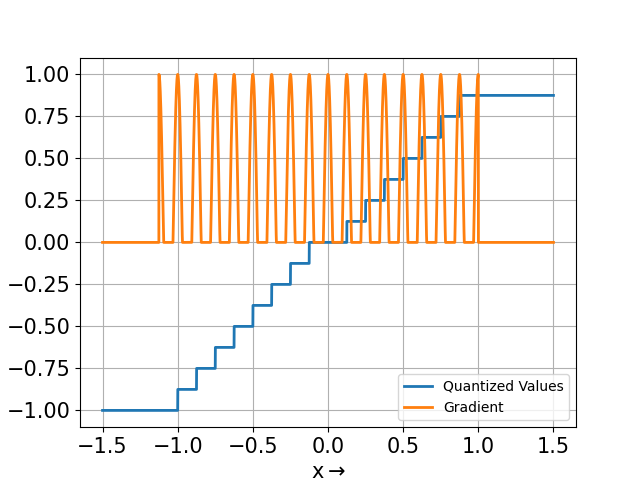}
\caption{Clipped Cosine Gradients for \\ Hidden State and Input Quantizations}
\label{fig:grads}
\end{subfigure}

\caption{Quantization of Sigmoid, Tanh, Hidden States and Inputs}
\label{fig:quants}
\end{figure*}

The NNA uses a symmetric linear quantization scheme to map 32-bit floating point numbers to 8-bit integers, rounded toward zero. Considering our previous example of $Q3.2$, the range of representable fractional  values is [-4, 3.75], which on the hardware is represented as integers in the range [-16, +15]. Any value less than -4 or greater than 3.75 is clipped to these limiting values, thus accumulating clipping error during static quantization. 
For dynamic quantization, the allowable dynamic scales are 1, 2, 4, and 16, which are calculated on the CPU and used to scale the original values into the quantizable range.  The outputs are uniformly quantized with equal step sizes and quantization intervals as illustrated in Fig. \ref{fig:grads}.

\subsection{Non-linear Activation Quantization}

The hyperbolic tangent and sigmoid functions are the commonly used non-linear activation functions in neural networks and are integral components in learning long range memory. They are expensive to compute even on CPUs, and thus are approximated on the NNA hardware through careful digital circuit design and error analysis. An efficient hardware implementation of activation functions is required to meet the performance, area, power and cost targets of neural accelerators. Multiple designs and approximation algorithms have been proposed to balance this trade-off \cite{5118213,4541553,4682175}. Usually, a combination of linear interpolators, shifting operations, look-up tables and multiplexers are used in the digital circuit to approximate the activation values. We are interested in translating the on-device operators into the model training workflow to train accelerator-aware models.

For the accelerator considered in this work, activation functions are implemented as a piece-wise linear approximator, which gives non-uniformly quantized 8-bit values as outputs. The design, modeling and analysis of these digital circuits are outside the scope of this work (please refer to \cite{5118213,4541553,4682175}). The approximation algorithm is optimized to balance the trade-off between accuracy and on-chip area, and leverages the fact that mathematically, the tanh and sigmoid functions are shifted and scaled versions of one another. It also reduces quantization error by carefully choosing non-uniform quantization centers to pack more quantization bins at parts of the functions with higher gradients. Despite the careful design, it is not possible to circumvent the information loss due to quantization. Given that the activation functions play a key role in learning long-term memory, as we show later in the results, switching from high precision to 8-bit values leads to a large performance degradation. To alleviate this performance degradation, these values need to be computed on the CPU, which requires constant transfer of data on the chip to the CPU in every time step of the LSTM cell to perform activation functions, and back to the NNA to perform MAC operations, incurring additional processing latency. A major contribution of this paper is to incorporate the bit-exact hardware operators into the training workflow with meaningful gradient backpropagation and further fine-tune the model to non-uniformly quantized activations, yielding 5-7\% latency reduction and negligible, less than 1\% relative WER degradation.

\section{Accelerator-Aware Training}
\label{sec:AAT}
We propose a two-stage method to address the errors propagated by the activation and intermediate quantizations in the model. Before diving into the implementation details, we first discuss the motivation for choosing such an approach.

The sigmoid and tanh functions saturate approximately at +/-7  and +/-4 respectively (Figs. \ref{fig:sigmoid} and \ref{fig:tanh}). However, in the process of quantizing the activations, a large clipping and precision error is accumulated between +/-7  and +/-4, for the activations when compared to the FP-32 (full precision) values. This is because a significant portion of the inputs to the activation functions in the encoder and decoder LSTMs of a fully trained RNN-T model lie in this error-prone range. Due to the large number of values present in the saturation range of these functions, the vanishing gradient problem hinders learning using backpropagated gradients, which is already affected due to the addition of quantization. To overcome this issue, we add an activity regularization loss on the inputs to the activation functions as described in Section \ref{ssec:act_reg}.

For tuning the inputs, hidden states and activation function values into the quantization levels, we perform a bit-exact quantization in the forward pass and approximate the gradients in the backward pass using straight-through estimation. This is detailed for the various components in Section \ref{ssec:act_quant}.

\subsection{Activity Regularization}
\label{ssec:act_reg}

In Stage I of training, we initialize the model with random weights and train it from scratch with an activity regularization loss on the outputs $z_0$, $z_1$, $z_2$, and $z_3$ of the LSTM cell \cite{hochreiter1997long} to alleviate the quantization errors as discussed above. The additional loss term aims to restrict the range of the inputs to the activations functions by penalizing the values proportional to their distance outside the allowable range $[z_{min}, z_{max}]$. We achieve this by using a regularization loss, $L_{activity}$, defined by a shifted ReLU function.
\begin{equation}
L_{activity} = ReLU(z + z_{min}) + ReLU(z - z_{max})
\end{equation}
\vspace{-0.3cm}
\begin{equation}
\label{eq:tot_loss}
L_{total} = L_{model} + \lambda L_{activity}
\end{equation}
where the Rectified Linear Unit, $ReLU(x)$ = $max(0, x)$ and $z$ is the concatenated array [$z_0$, $z_1$, $z_2$, $z_3$]. For values of $z$ outside the allowable range, the loss adds a proportional penalizing term, thus restricting their values. The loss is differentiable with respect to the input except at the range extrema. However, the derivative at these extrema is approximated to be 0, making it a loss function compatible for backpropagation. We train the model with $L_{activity}$ added as an additional term in the total loss $L_{total}$ (see Eq. \ref{eq:tot_loss}), where $\lambda$ is used to weigh the activity regularization term.

\subsection{Quantization of Inputs, Hidden States and Activations}
\label{ssec:act_quant}

In Stage II of training, we initialize the model with the trained weights from stage I and apply accelerator-aware training for the linear quantization of inputs and hidden states, and the non-linear quantization of the tanh and sigmoid activation functions. 
\subsubsection{Bit-exact Quantization Replication during Forward Pass}

\begin{algorithm}
	\caption{QAT for Inputs, Hidden States and Activations}
	\textbf{For Inputs and Hidden States}
	\label{algo:qat}
		 \begin{algorithmic}
		\STATE  \nonumber \emph{Forward pass:} Bit-exact linear quantization of input tensor X\\
		\nonumber  \hspace{0.2in}   a. $Y$ = $\mathcal{Q}$(X)\\
		\STATE \nonumber  \emph{Backward pass:} Clipped Cosine Gradients\\
		\nonumber  \hspace{0.1in}  a. Frequency,  $f_q$ = $1$\\
		\nonumber  \hspace{0.1in}  b. $\frac{\partial Y}{\partial X} $ = clip$(\cos(2\pi f_q X), min=0, max=1)$ \\ \COMMENT{Gradients are clipped to 0 at the quantization bin mid-points}\\
		\end{algorithmic}
		\textbf{For Non-linear Activations}
    \begin{algorithmic}
		\STATE  \nonumber \emph{Forward pass:} Bit-exact non-linear quantization of input tensor X\\
		\nonumber  \hspace{0.1in}   a. $Y$ = $\mathcal{Q}_{act}$(X)\\
		\STATE \nonumber \emph{Backward pass:} tanh and sigmoid gradients\\
		\nonumber  \hspace{0.1in}  a. For tanh,  $\frac{\partial Y}{\partial X} $ = $1 - tanh^2(X)$\\
		\nonumber  \hspace{0.1in}  b. For sigmoid,  $\frac{\partial Y}{\partial X} $ = $sigmoid(X) * (1 - sigmoid(X))$\\
	\end{algorithmic}
\end{algorithm}

\begin{itemize}
\item For the linear quantization of inputs and hidden states, we use Eq. \ref{eq:quantdequant} in the forward pass. 
\item For the non-linear activation functions, the on-device linear interpolator in Fig. \ref{fig:quants} is replicated during the forward pass. We denote this bit-exact quantization function implemented in the training workflow as $\mathcal{Q}_{act}$ (see Algorithm \ref{algo:qat}). Although we demonstrate it here for the $sigmoid$ and $tanh$ functions for a particular hardware implementation, this method can be generally applied to any other similar linearly or non-linearly quantized functions. 
\end{itemize}

\subsubsection{Backpropagation Using Meaningful Gradients}
Since quantization is not a differentiable process, we employ the straight-through estimation method \cite{bengio2013estimating} for the forward pass and backpropagate meaningful gradients through the various quantization nodes in the backward pass.
\begin{itemize}
\item For the inputs and hidden states, we use clipped cosine gradients as illustrated in Fig. \ref{fig:grads}. Instead of passing a unity gradient throughout the range of the inputs at the quantization node as proposed in \cite{bengio2013estimating}, we propose to backpropagate through the quantization node a periodic clipped cosine gradient to disincentivize values to occupy unwanted areas that lie outside the quantized levels.  The clipped cosine gradients drive the values into surrounding quantization bins, while leaving the values around the center of the bins unchanged.
\item For the activations, we use the full precision $sigmoid$ and $tanh$ gradients.
\end{itemize}


\section{Experimental Setup}
\label{sec:setup}
We conduct our experiments under two setups A and B. In setup A, we train a small RNN-T model with 5 layers of an encoder and 2 layers of a decoder with an additive and feedforward joint network. The number of hidden units in the LSTM cell for setup A is 640, yielding a model with~26M parameters. Setup B is a larger RNN-T variant with the same number of encoder, decoder and joint network layers, in which the LSTMs have 1024 hidden units, yielding a model with ~66M parameters. For both setups, the baseline models are trained for a total of 600k steps,  with 5k steps of warmup to a learning rate (LR) of 5e-4 which was held constant for 150k steps, followed by an exponential decay to a learning rate of 1e-5 for the remaining steps. The Accelerator-aware training (AAT) models are trained in Stage I for 500k steps with $\lambda=2$ (see Eq. \ref{eq:tot_loss}), and the same LR schedule as the baseline. This was followed by 5k-10k steps of training in Stage II with a constant learning rate of 1e-4.  The acoustic features are 64-dimensional Log Mel Filterbank Energies (LFBE) with a window size of 25 ms and 10 ms overlap between frames.

\vspace{0.2cm}
All models are trained with absolute cosine quantization aware regularization for 8-bit on-device quantization of weights \cite{NguyenAle2020}. To train models under both setups, we use an in-house collection of a far-field English training dataset with 270k hours of audio. For evaluation, we use 6 test sets: 3 for Setup A, and 3 for Setup B. Since Setup A has a smaller model, we evaluate it on a smaller number of intents. The number of utterances in each test set is available in Table \ref{table:wer}.

\section{Results and Discussion}
\label{sec:results}
In Table 1, we compare the relative WER reduction (WERR) performance between the baseline and AAT models. Note that a positive WERR number indicates a WER improvement, whereas negative numbers signify WER degradation. Furthermore, in order to evaluate the effectiveness of each proposed stage, we present WER  for each stage in Table 1.1. Here, Q denotes quantized WER numbers. Model A-I and A-2 represent the model in Setup A after Stage 1 and Stage II trainings, respectively. All WERR numbers are computed between the quantized (Q) version of the respective model and the un-quantized version of the baseline model. For example, the WERR of quantized Model A-II with respect to the un-quantized Baseline-A is +0.3\%.

\begin{table}[ht]
\centering
\captionsetup{width=1.0\linewidth}
\caption{Relative WER Reductions (WERR) for Setup A and B. WERR numbers are computed between the quantized (Q) version of the respective model and the un-quantized version of the baseline (i.e. Baseline (UN-Q) = 0.0).}
\label{table:wer}
\textbf{Table 1.1: Setup A}\\
\vspace{0.1cm}
\begin{tabular}{c|c|c|c}
\hline
  \multirow{1}{*}{Datasets} &    \multicolumn{3}{c}{WERR (\%)} \\
 \cline{2-4}
(Num. Utts)& \multicolumn{1}{c|}{Baseline (Q)} &  \multicolumn{1}{c|}{A-I (Q)} &  \multicolumn{1}{c}{A-II (Q)}\\
 \hline
 A-D1 (42688) &  -9.9&  -5.8  & +0.3  \\
 A-D2 (26226) &  -6.0&  -3.0    & +0.4  \\
 A-D3 (46208)  &  -8.0 &  -3.9 & +0.8   \\

\hline
\textbf{Stage I}& \multicolumn{1}{c|}{\xmark} & \multicolumn{1}{c|}{\cmark} & \multicolumn{1}{c}{\cmark} \\
\textbf{Stage II}& \multicolumn{1}{c|}{\xmark} & \multicolumn{1}{c|}{\xmark} & \multicolumn{1}{c}{\cmark} \\
\hline
\hline
\end{tabular}

\vspace{0.7cm}
\textbf{Table 1.2: Setup B}\\
\vspace{0.1cm}
\begin{tabular}{c|c|c}
\hline
  \multirow{1}{*}{Datasets} &    \multicolumn{2}{c}{WERR (\%)} \\
  \cline{2-3}
 (Num. Utts) & \multicolumn{1}{c|}{Baseline (Q)} &  \multicolumn{1}{c}{B-I (Q)} \\
  
 \hline
 B-D1 (155936) & -5.4 &  -0.4 \\
 B-D2 (46530) & -7.2 &  -0.6   \\
 B-D3 (20279) &  -3.0&  +1.3  \\

\hline
\textbf{Stage I}& \multicolumn{1}{c|}{\xmark} & \multicolumn{1}{c}{\cmark} \\
\textbf{Stage II}& \multicolumn{1}{c|}{\xmark} & \multicolumn{1}{c}{\cmark}\\
\hline
\hline
\end{tabular}

\end{table}

As shown in the Table \ref{table:wer}, quantizing the inputs, hidden states and activations for the baseline models without the proposed AAT approach leads to 6-10\% relative WER degradation for Setup A, and 3-7\% for Setup B across our testing datasets. This degradation is reduced to be within 1\% relative to the baseline performance after the two-stage AAT.  In particular, as expected, we observed a larger degradation for the model in Setup A than the model in Setup B since the former has smaller number of parameters and is more susceptible to quantization than the latter. It is observed that by using the two-stage approach, we can effectively reduce the WER gap between quantized and unquantized versions for small and large models alike.

\begin{table}[ht]
\label{table:latency}
\centering
\captionsetup{width=1.0\linewidth}
\caption{Normalized Latency Measurements for Setup A and B. All values are normalized with respect to the baseline latencies at p50, for the respective setups.}
\begin{tabular}{c|c|cc|cc}
\hline
 \multirow{2}{*}{Setups}&\multirow{2}{*}{Statistic} & \multicolumn{2}{c|}{Baseline} &  \multicolumn{2}{c}{AAT}\\
&& \textbf{EL} &  \textbf{UPL} &  \textbf{EL} &   \textbf{UPL}  \\
 \cline{1-6}
 \multirow{3}{*}{Setup A} & \textbf{p50} & 1.0 & 1.0  & 0.94 &0.95 \\
& \textbf{p90} & 1.17 & 1.49 & 1.11 &1.35 \\
& \textbf{p99} & 1.34  & 7.21 & 1.28 &7.21 \\
 \cline{2-6}
\multirow{3}{*}{Setup B} & \textbf{p50} & 1.0 & 1.0 & 0.94 &0.94\\
& \textbf{p90} & 1.53& 1.43 & 1.42& 1.35\\
& \textbf{p99} & 2.41 & 1.90 & 2.16 &2.36\\
\hline
\hline
\end{tabular}
\end{table}

\vspace{0.2cm}
To demonstrate the latency gains, we provide the ASR engine latency (EL) and UPL numbers in Table 2. Here, EL measures the time elapsed between the user completing the utterance and the ASR recognition result being available in the ASR engine. UPL denotes the time elapsed between the user completing the utterance, and when the system responds with an appropriate dialogue / action. Thus, UPL includes EL plus the other server and device side processing required to fulfill the user's request. We conduct on-device tests with 500 utterances for  both baseline and AAT models in setups A and B, and report both EL and UPL numbers at the 50th, 90th and 99th percentiles. As before, positive relative numbers signify reductions in latency whereas negative  numbers signify degradation. We see that by compiling the AAT models, we can get approximately 4 - 9\% EL and UPL reductions at p50. AAT enables this by allowing us to shift to a more quantized, faster NNA data path during inference with negligible degradation in performance. 

\section{Conclusions}
\label{sec:concs}

In this work, we introduce a two-stage AAT approach to alleviate the performance degradation due to performing post-training model conversion and quantized inference on the NNA. In particular, we incorporate the NNA quantization operators into model training and apply regularization as well as cosine gradients at the quantization nodes to reduce the performance gap between the in-training and post-training models. It is observed that with our proposed approach, there is little to no WER performance degradation between the unquantized baseline models and the quantized AAT counterparts. Compared to the original quantized baseline, the proposed two-stage quantized AAT models save 3-10\% relative WER degradation. Furthermore, AAT enables NPU/hardware execution of activation functions in runtime, leading to 5-7\% relative latency reductions. 




\bibliographystyle{IEEEbib}
\bibliography{strings,refs}

\end{document}